\documentclass{article}
\usepackage{spconf,amsmath,graphicx}
\usepackage{subfigure}
\usepackage{cite}
\usepackage{color}
\usepackage{hyperref}
\usepackage{breakurl}
\title{A Method for Identifying Origin of Digital Images \\ Using a Convolution Neural Network}
\name{Rong Huang\textsuperscript{1}, Fuming Fang\textsuperscript{2}, Huy H. Nguyen\textsuperscript{3}, Junichi Yamagishi\textsuperscript{2,3}, Isao Echizen\textsuperscript{2,3}\thanks{This work was partially supported by a JST CREST Grant (JPMJCR18A6, VoicePersonae project), Japan, and by MEXT KAKENHI Grants (16H06302, 17H04687, 18H04120, 18H04112, 18KT0051), Japan.}}
\address{\textit{\textsuperscript{1}College of Information Science and Technology, Donghua University, Shanghai, China} \\
\textit{\textsuperscript{2}National Institute of Informatics, Tokyo, Japan}\\
\textit{\textsuperscript{3}SOKENDAI (The Graduate University for Advanced Studies), Kanagawa, Japan}}
\begin{document}
\ninept
\maketitle
\begin{abstract}
The rapid development of deep learning techniques has created new challenges in identifying the origin of digital images because generative adversarial networks and variational autoencoders can create plausible digital images whose contents are not present in natural scenes. In this paper, we consider the origin that can be broken down into three categories: natural photographic image (NPI), computer generated graphic (CGG), and deep network generated image (DGI). A method is presented for effectively identifying the origin of digital images that is based on a convolutional neural network \linebreak (CNN) and uses a local-to-global framework to reduce training complexity. By feeding labeled data, the CNN is trained to predict the origin of local patches cropped from an image. The origin of the full-size image is then determined by majority voting. Unlike previous forensic methods, the CNN takes the raw pixels as input without the aid of ``residual map". Experimental results revealed that not only the high-frequency components but also the middle-frequency ones contribute to origin identification. The proposed method achieved up to 95.21\% identification accuracy and behaved robustly against several common post-processing operations including JPEG compression, scaling, geometric transformation, and contrast stretching. The quantitative results demonstrate that the proposed method is more effective than handcrafted feature-based methods.
\end{abstract}
\begin{keywords}
image origin, convolutional neural network, local-to-global, robustness
\end{keywords}
\vspace{-0.06cm}
\section{Introduction}
\vspace{-0.06cm}
\label{sec:intro}

\begin{figure}[t]
    \centering

    \subfigure[NPI]{\includegraphics[height=3.2cm]{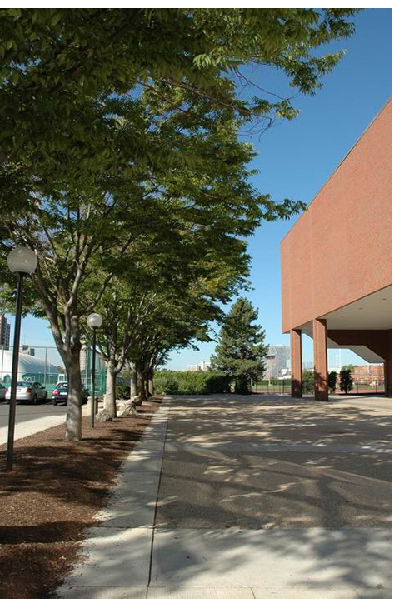}}
    \hspace{-0.1cm}
    \subfigure[CGG]{\includegraphics[height=3.2cm]{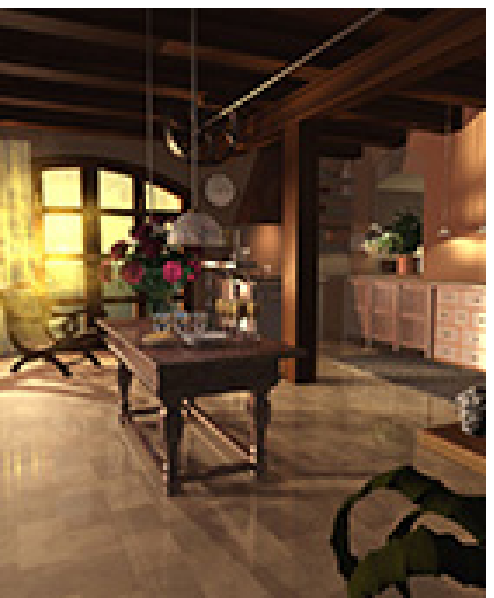}}
    \hspace{-0.1cm}
    \subfigure[DGI]{\includegraphics[height=3.2cm]{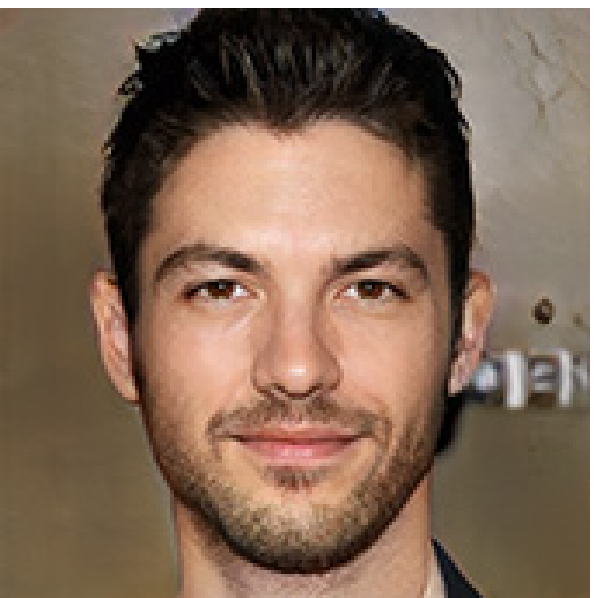}}

    \vspace{-0.3cm}
    \caption{Example digital images with different origins.}
    \label{fig:example}
\vspace{-0.5cm}
\end{figure}

Nowadays, with the proliferation of powerful computer graphics tools, such as Softimage XSI, Maya, and TerraGen, it has become easy for non-professional technicians to artificially create digital scenes without leaving any perceptible clues. Figures \ref{fig:example} (a) and (b) show a natural photographic image (NPI) and a computer generated graphic (CGG), respectively. At first glance, it is not obvious that the second image was computer generated. In media forensics, substantial efforts have been made to distinguish between a CGG \linebreak and an NPI. Statistical clues based on wavelet coefficients \cite{Lyu_TSP}, histograms \cite{Wu_ICIP}, edge pixels \cite{Zhang_IWDW}, texture \cite{Wang_CVIU}, entropy \cite{Nguyen_IWDW}, and multi-fractal spectrum \cite{Peng_IJEC} have been used to design handcrafted features geared towards a particular classifier. Furthermore, pixel-level inconsistencies such as JPEG compression artifacts \cite{Bianchi_TIFS}, demosaicking clues \cite{Ferrara_TIFS}, lens aberration \cite{Yerushalmy_IJCV}, and sensor pattern noise \cite{Chen_TIFS} are useful indicators for identification. Recently, deep neural networks have been the workhorse for a wide variety of computer vision tasks, including image classification, image annotation, and object detection. Driven by data, CNNs have the ability to automatically learn hierarchical representations, and thereby achieve better generalization in an end-to-end manner. Hence, CNN-based methods \cite{Rahmouni_WIFS,Yu_ICIP,Quan_TIFS} have been quickly adopted by the multimedia security community for distinguishing CGGs from NPIs.

Deep neural networks, however, are like a double-edged sword: they can be used not only for identifying origin but also for generating photorealistic images if the network is structured as a generative adversarial network (GAN) or variational autoencoder (VAE). Since most generative models can be trained in an unsupervised manner, it is not prohibitively expensive for an attacker to generate fake scenes that can be maliciously used for illegal purposes. As evidenced by the deep network generated image (DGI) in Fig.\ref{fig:example} (c), it is not trivial to identify DGIs with the naked eye. Several countermeasures have been proposed recently. McCloskey and Albright \cite{McCloskey} found that the frequency of saturated and under-exposed pixels is suppressed by the generator's normalization steps, which provide useful discriminative traces that enable a DGI to be distinguished from an NPI. Marra {\em et al.} \cite{Marra} revealed that specific correlation exists in the noise residual of a DGI, which can be viewed as evidence of the origin of digital images. Li {\em et al.} \cite{Li_color} exploited the disparities in color components and designed a 588-dimensional feature vector based on multiple color co-occurrence matrices. In addition, some CNN-based methods \cite{Cozzolino,Yu,Meso,BTAS} attempt to identify DGIs and/or localize fake regions for facial images.

According to the above survey, we conclude that digital images are created by one of three ways: (i) being captured with a CCD or CMOS sensor, which counts the photons passing through a color filter array, (ii) being synthesized using model-based rendering software, and (iii) being generated using data-driven deep learning model (e.g., GAN and VAE). However, the existing methods based on handcrafted features \cite{Lyu_TSP,Wu_ICIP,Zhang_IWDW,Wang_CVIU,Nguyen_IWDW,Peng_IJEC,Bianchi_TIFS,Ferrara_TIFS,Yerushalmy_IJCV,Chen_TIFS,McCloskey,Marra,Li_color} can only distinguish a CGG \linebreak or DGI from an NPI. Therefore, an end-to-end method that can automatically probe discriminative features and complete the origin identification task is needed to ensure the information security of digital images. To achieve this, we develop a CNN-based method in a local-to-global framework. The experimental results demonstrate that the proposed method is better in terms of identification accuracy and robustness compared with handcrafted feature-based methods.

\begin{figure}[t]
    \centering
    \begin{tabular}{c}
        \scalebox{0.2}[0.2]{\includegraphics{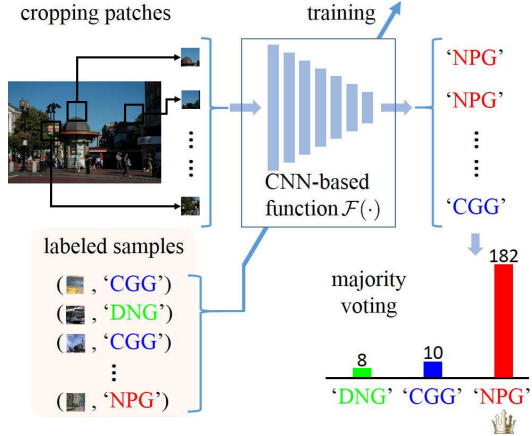}}
    \end{tabular}
    \vspace{-0.3cm}
    \caption{Flowchart of proposed method.}
    \label{fig:fm}
\vspace{-0.5cm}
\end{figure}

\vspace{-0.06cm}
\section{Proposed Method}
\vspace{-0.06cm}
\label{sec:format}

The identification task considered in this paper is a three-class problem. Suppose that a training set comprises $N$ samples along with their labels $\{(x_1,y_1),$ $(x_2,y_2),$ $\cdots,$ $(x_n,y_n),\cdots,(x_N,y_N)\}$, where $n=1,2,\cdots,N$. The notation $x_n$ represents the $n^{\text{th}}$ to-be-identified sample while $y_n$ corresponds to its label. Our goal is to exploit the training set and establish a CNN-based mapping function $\mathcal{F}(\cdot)$ so as to make correct predictions for new samples.
\par We use a local-to-global framework to reduce training complexity. Specifically, for an RGB color image, we crop $M$ local patches, each of size $224\times 224 \times 3$. After the CNN model is trained, it is expected to predict the correct label, denoted by $\hat{y}_n$, for each local patch. The origin of the full-size image is then determined by majority voting. Figure \ref{fig:fm} schematically illustrates the flowchart of the proposed method. Using the local-to-global framework not only reduces training complexity but also forces the CNN model to learn forensic clues rather than recognize the image contents.

\vspace{-0.06cm}
\subsection{Patch Cropping}
\vspace{-0.06cm}
Sampling representative data is an important prerequisite for training a statistical learning model. Motivated by the observation that highly textured areas usually contain clues about factitiousness, we preferentially crop patches containing rich edge pixels. Specifically, the edge pixels are  detected by Canny detector. The high and low threshold values are set to 50 and 100, respectively. The image is then partitioned into overlapping patches with an appropriate stride to ensure that the number of candidate patches exceeds $20M$. The number of edge pixels within a candidate patch is considered to be the fitness value. From the $20M$ candidate patches, roulette wheel selection, which is widely used in genetic algorithms, is performed to choose $M$ ones without replacement. Compared with random cropping, the roulette-based strategy can avoid cropping too many smooth patches.

\vspace{-0.06cm}
\subsection{Data Augmentation}
\vspace{-0.06cm}
As a regularization technique, data augmentation protects well \linebreak against overfitting. The training set is artificially expanded by generating new instances from existing ones. To this end, the patches are rotated by $90^{\circ}$, $180^{\circ}$, and $270^{\circ}$, and horizontal and vertical flipping are performed. A total of $6M$ patches can be harvested from a full-size image by combining these geometric transformations.
\par Furthermore, these newly generated patches help the model to be more robust against rotation and flipping. The experimental results presented in Section \ref{exp} support this claim.

\begin{figure}[t]
    \centering
    \begin{tabular}{c}
        \scalebox{0.2}[0.2]{\includegraphics{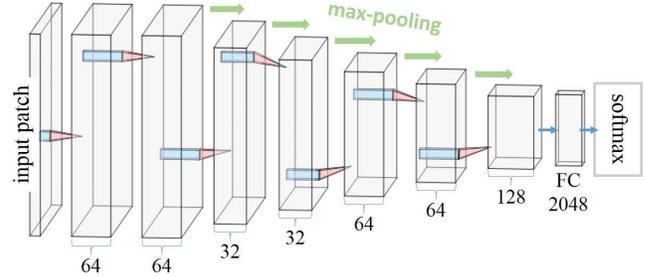}}
    \end{tabular}
    \vspace{-0.3cm}
    \caption{Proposed CNN architecture.}
    \label{fig:nn_arch}
\vspace{-0.5cm}
\end{figure}

\vspace{-0.06cm}
\subsection{Network Architecture}
\vspace{-0.06cm}
\label{sec:typestyle}
CNNs typically consist of a cascade connection of multiple convolutional layers, pooling operations, and one or more fully connected (FC) layers. Each convolutional layer, which can be viewed as a group of filters, is trained to produce an appropriate feature representation. Each pooling operation, which computes a summary statistic of a certain layer's output, helps to make the feature representation approximately invariant to small translations. The learned hierarchical representation is fed into the FC layer, which plays a role in making correct predictions.
\par The proposed CNN architecture is shown in Fig. \ref{fig:nn_arch}. The input local patches are $224\times 224 \times 3$. The CNN model consists of seven convolutional layers, five pooling operations and one FC layer. Specifically, there are 64, 64, 32, 32, 64, 64, and 128 filters for the 7 convolutional layers, respectively. The filters are all $3\times 3$ and have a unit stride. A rectified linear unit (ReLU) is used for activation. In each convolutional layer, batch normalization is used to mitigate the problem of internal covariate shift, making the model significantly easier to train. The outputs of the last five convolutional layers are further processed by a $2\times 2$ max-pooling operation with stride 2. The final FC layer contains 2048 neural nodes followed by a three-way softmax function.
\par In previous steganalysis and forensics methods \cite{Li_color,Cozzolino}, the input is pre-processed using fixed high-pass filters to suppress image contents and extract the so-called ``residual map". However, there is no evidence to suggest that the middle (or low) frequency components make no contribution to the forensics task. We therefore skip this pre-processing step and make all the filters trainable in the hope of adaptively probing useful information.

\vspace{-0.06cm}
\subsection{Training}
\vspace{-0.06cm}
\label{sec:majhead}

We trained the CNN model using two scenarios. In scenario 1, the first two convolutional layers simply duplicate the previously trained `conv1\_1' and `conv1\_2' of the VGG19 model \cite{vgg}, and the corresponding weights are kept fixed during future training. In scenario 2, all layers are trained without any constraints by directly using task-dependent data. For convenience, we call these two scenarios `vgg' and `ada', respectively.

\par The cross-entropy between $y_n$ and $\hat{y}_n$, where $n=1,2,\cdots,N$, is defined as the cost function. A gradient-based stochastic optimization method called ADAM \cite{adam} is used to update the weights with a learning rate of $10^{-4}$, a minibatch size of 128, and three default settings ($\beta_1=0.9$, $\beta_2=0.999$, and $\epsilon=10^{-8}$). To prevent the network from overfitting, a powerful regularization technique called Dropout \cite{dropout} is applied during the training phase to the FC layer by setting the retention probability to 0.2. We empirically stop the training process after ten epochs.

\vspace{-0.06cm}
\subsection{Dataset}
\vspace{-0.06cm}
\label{ssec:subhead}
\par We used a standard dataset called Columbia Photographic Images and Photorealistic Computer Graphics (PRCG) \cite{dataset1}, which contains 800 CGGs collected from the Internet and 800 NPIs obtained from the personal collections. To prepare DGIs, we used a pre-trained progressive GAN (ProGAN) \cite{proGAN} to generate a sufficient number of high-quality images\footnote{The ProGAN was trained on the LSUN dataset \cite{dataset2} (category-wise) and CelebA-HQ dataset \cite{dataset3}.}. Then, we manually screened out the 800 most plausible ones and grouped them into the third set. Unlike methods that focus on only facial images \cite{Cozzolino,Yu,Meso,BTAS}, our proposed method aims at origin identification. Therefore, the images in the dataset encompass various scenes and contents including human faces, animals, landscapes, objects, \linebreak plants, buildings, vehicles, etc., for each class (NPI, CGG, and DGI). This enables us to train the model and thus make it more suitable for real-world application. We show some samples of the dataset online \url{https://nii-yamagishilab.github.io/Samples-Rong/Attack-Type-detection/}.

\vspace{-0.06cm}
\section{Evaluation}
\vspace{-0.06cm}
\label{exp}
We experimentally evaluated our proposed method: (i) to determine its identification accuracy, (ii) to study which frequency components are more useful for origin identification, and (iii) to determine its \linebreak robustness against several post-processing operations including JPEG compression, scaling, geometric transformation, and contrast stretching. For comparative studies, we used three handcrafted feature-based methods \cite{Wu_ICIP,Li_color,McCloskey} as baselines.

\vspace{-0.06cm}
\subsection{Evaluation Method}
\vspace{-0.06cm}
In data preparation, we randomly divided the image dataset into three subsets: training (70\%), validation (10\%), and test (20\%). From each training image, we cropped $M=200$ local patches each of size $224\times224\times3$ and then performed geometric transformations to expand the training set. This produced $N=2.016\times 10^6$ training samples. The CNN model was implemented using the TensorFlow framework. The computing platform consists of an i7-6770K 4GHz CPU and a NVIDIA GeForce GTX 1070 GPU with 16GB memory.
\par For the three baselines, we extracted the tailor-made features and uniformly used a discriminant analysis classifier. For the histogram-based method \cite{Wu_ICIP}, the number of bins was set to seven, as recommended by the authors. We simply calculated the average accuracy over three color channels. For the saturation-based method \cite{McCloskey}, each image was first converted from RGB into grayscale, and then an 8-dimensional descriptor was constructed by combining both the under- and over-exposed pixel frequencies. For the color co-occurrence-based method \cite{Li_color}, the order of the co-occurrence matrix was set to 3 under a sample-aware scenario.

\begin{figure}[t]
    \centering

    \subfigure{\includegraphics[height=2.45cm]{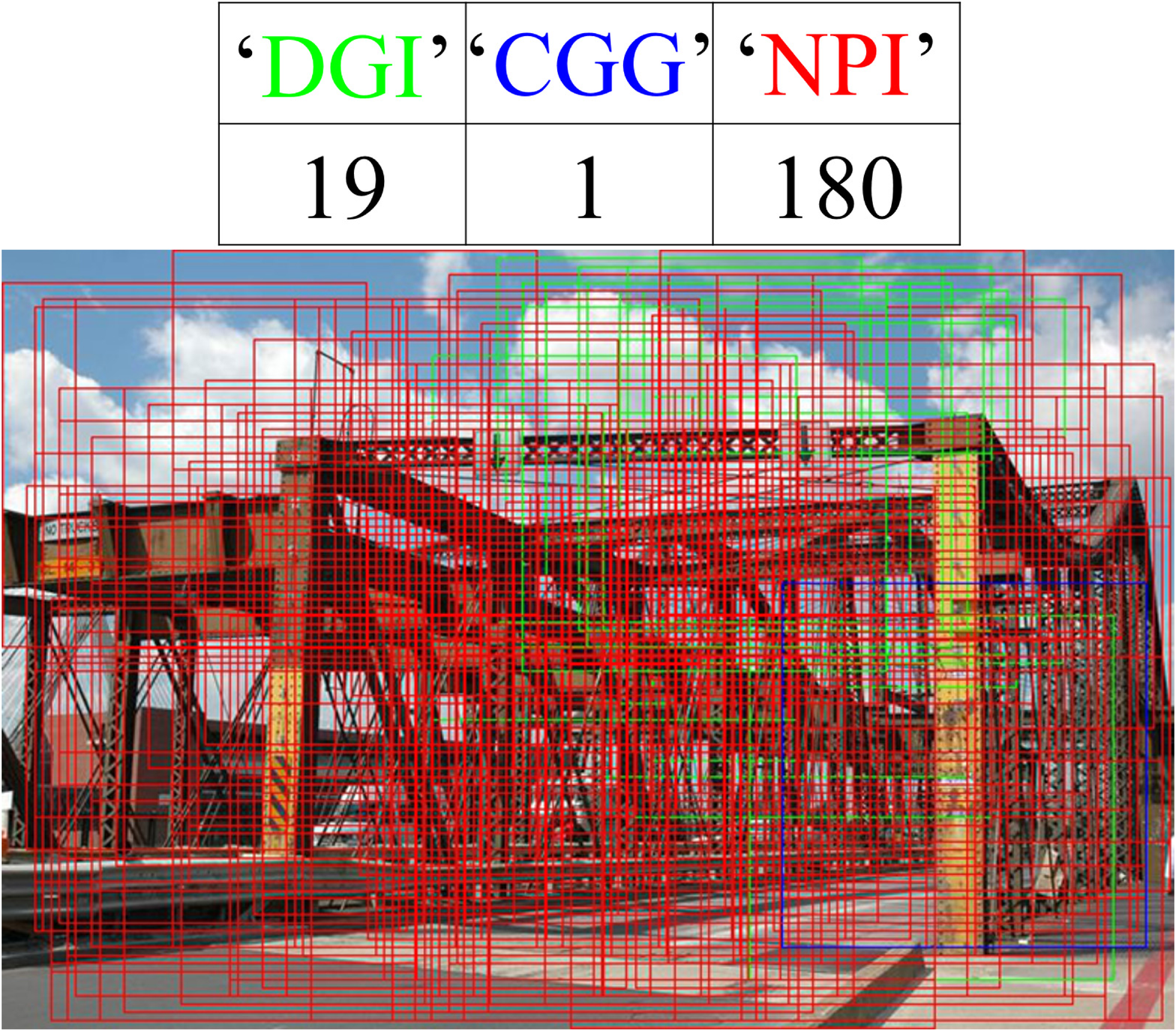}}
    \hspace{-0.1cm}
    \subfigure{\includegraphics[height=2.45cm]{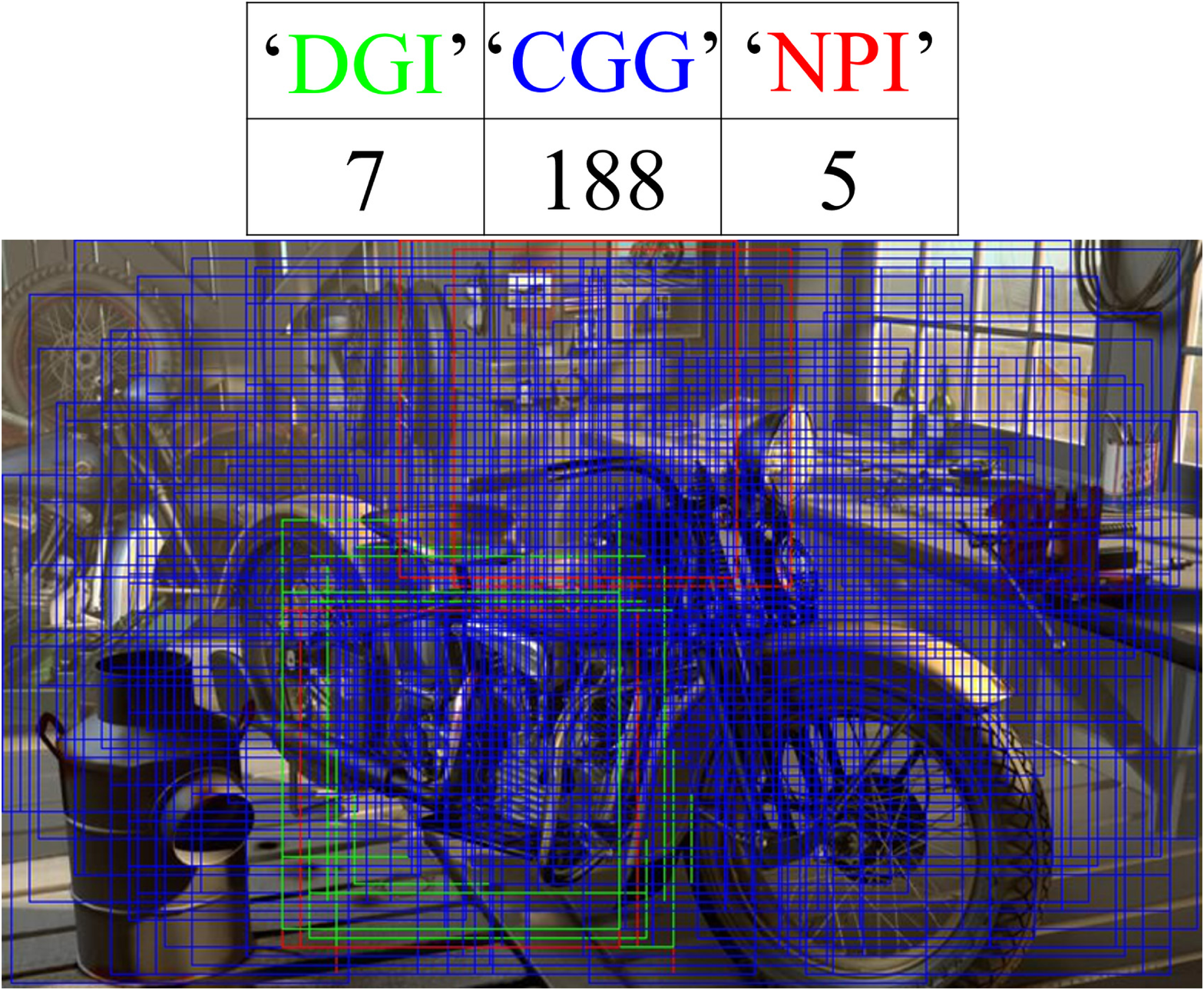}}
    \hspace{-0.1cm}
    \subfigure{\includegraphics[height=2.45cm]{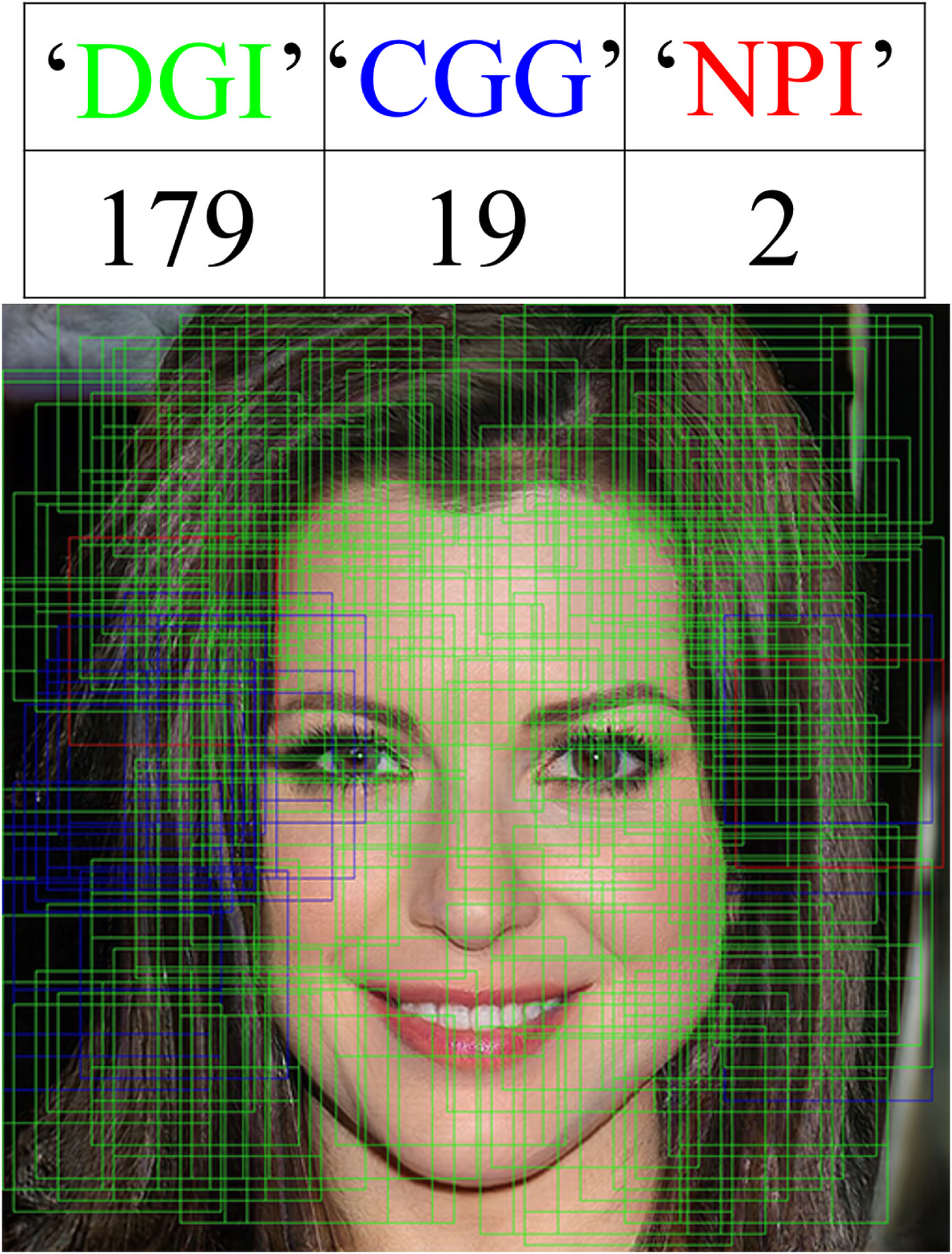}}\addtocounter{subfigure}{-3}

    \vspace{-0.2cm}

    \subfigure[NPI]{\includegraphics[height=2.45cm]{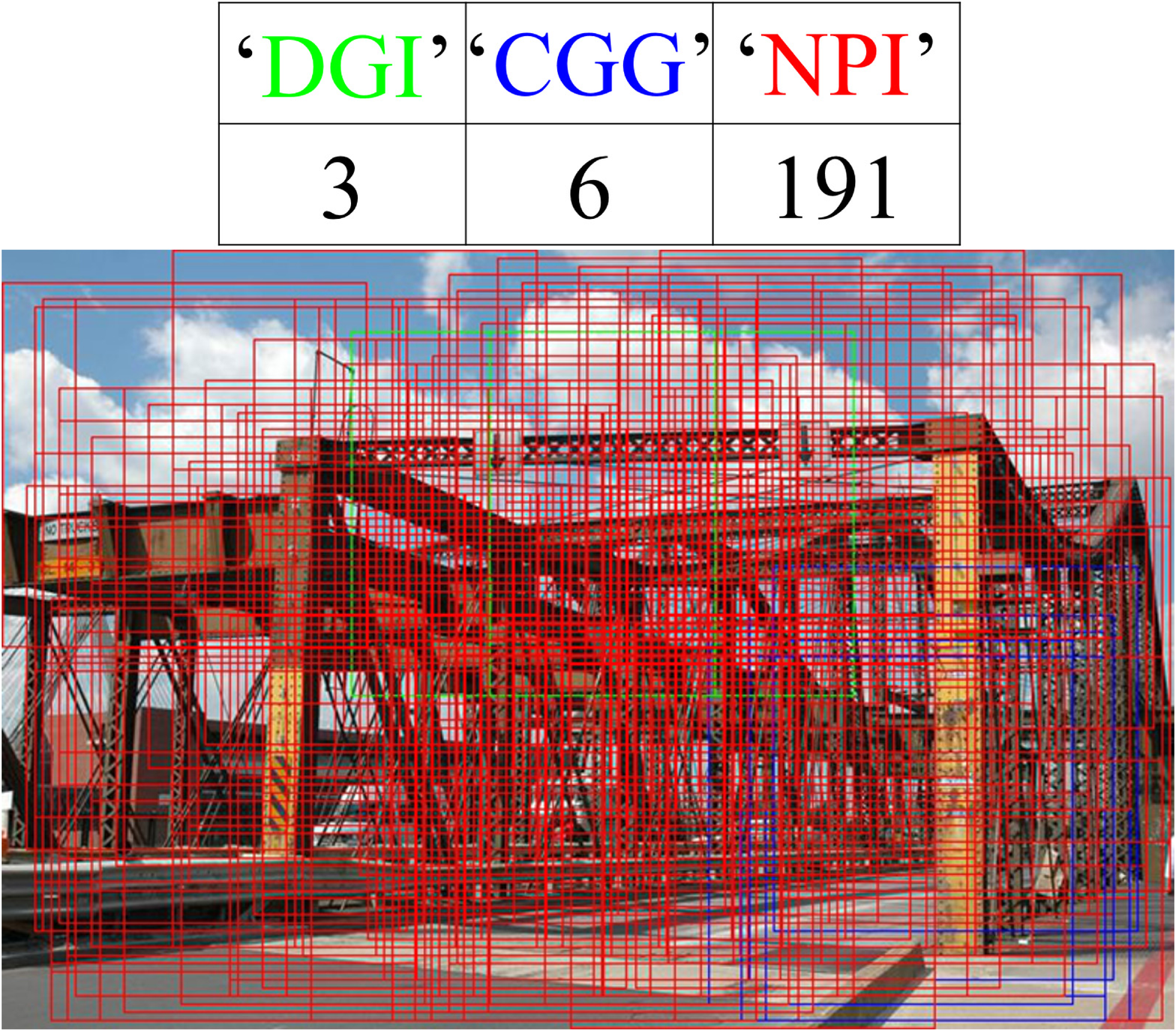}}
    \hspace{-0.1cm}
    \subfigure[CGG]{\includegraphics[height=2.45cm]{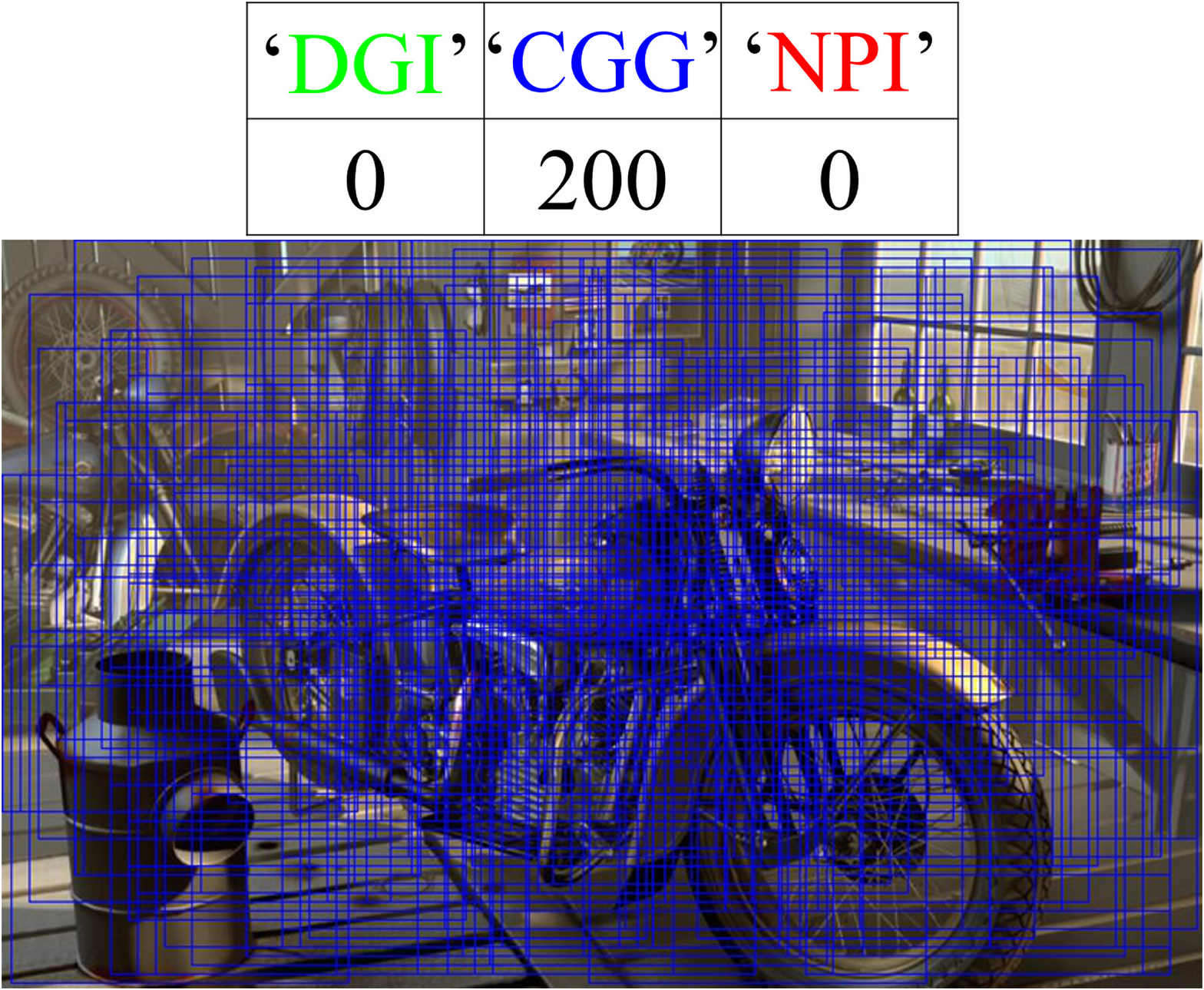}}
    \hspace{-0.1cm}
    \subfigure[DGI]{\includegraphics[height=2.45cm]{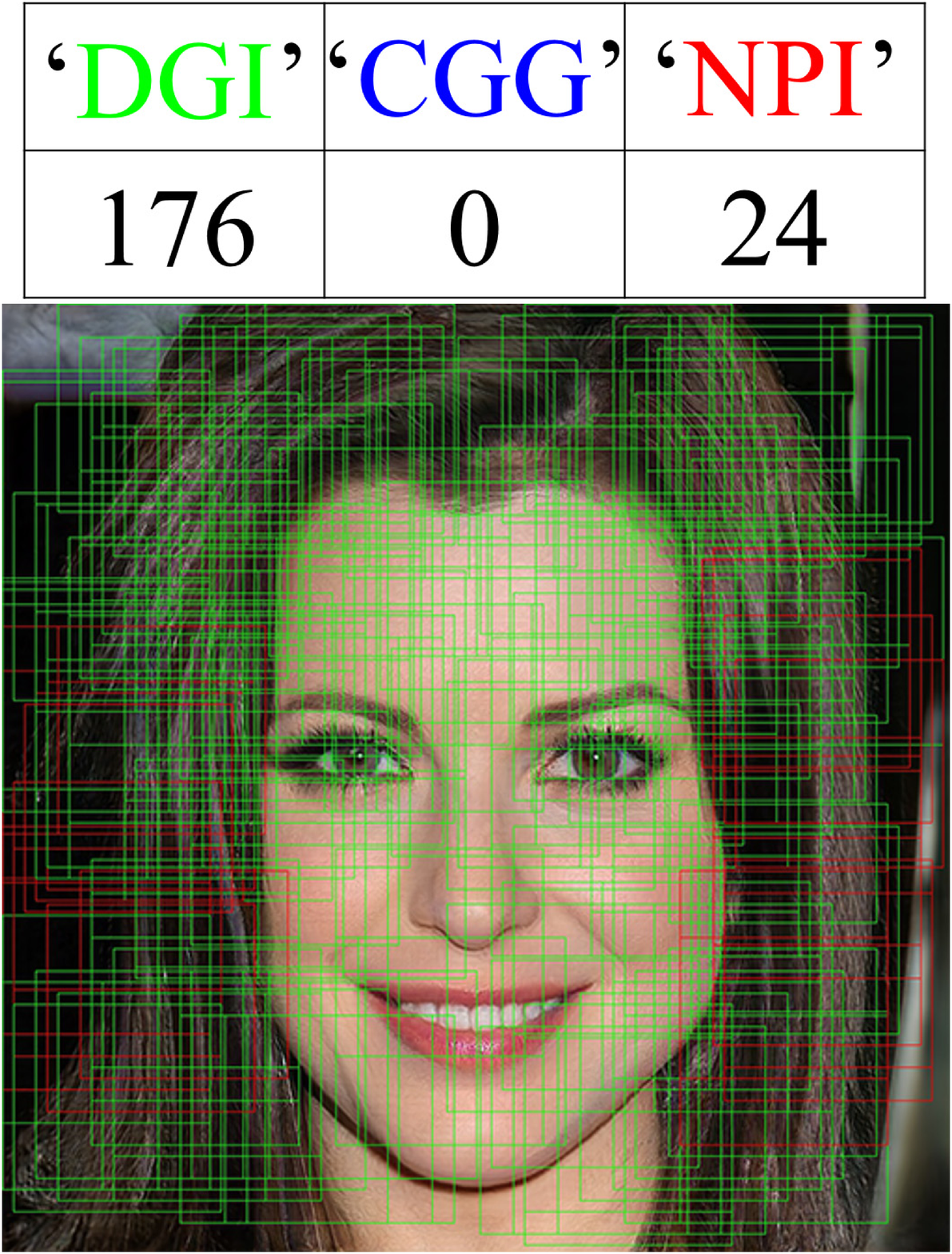}}

    \vspace{-0.3cm}
    \caption{Resulting examples of origin predications at patch level and quantitative results of majority voting. First row shows results for `vgg' scenario; second row shows results for `ada' scenario.}
    \label{fig:patch_r}
\vspace{-0.4cm}
\end{figure}

\begin{figure*}[t]
    \centering

    \subfigure[JPEG]{\includegraphics[height=3.28cm]{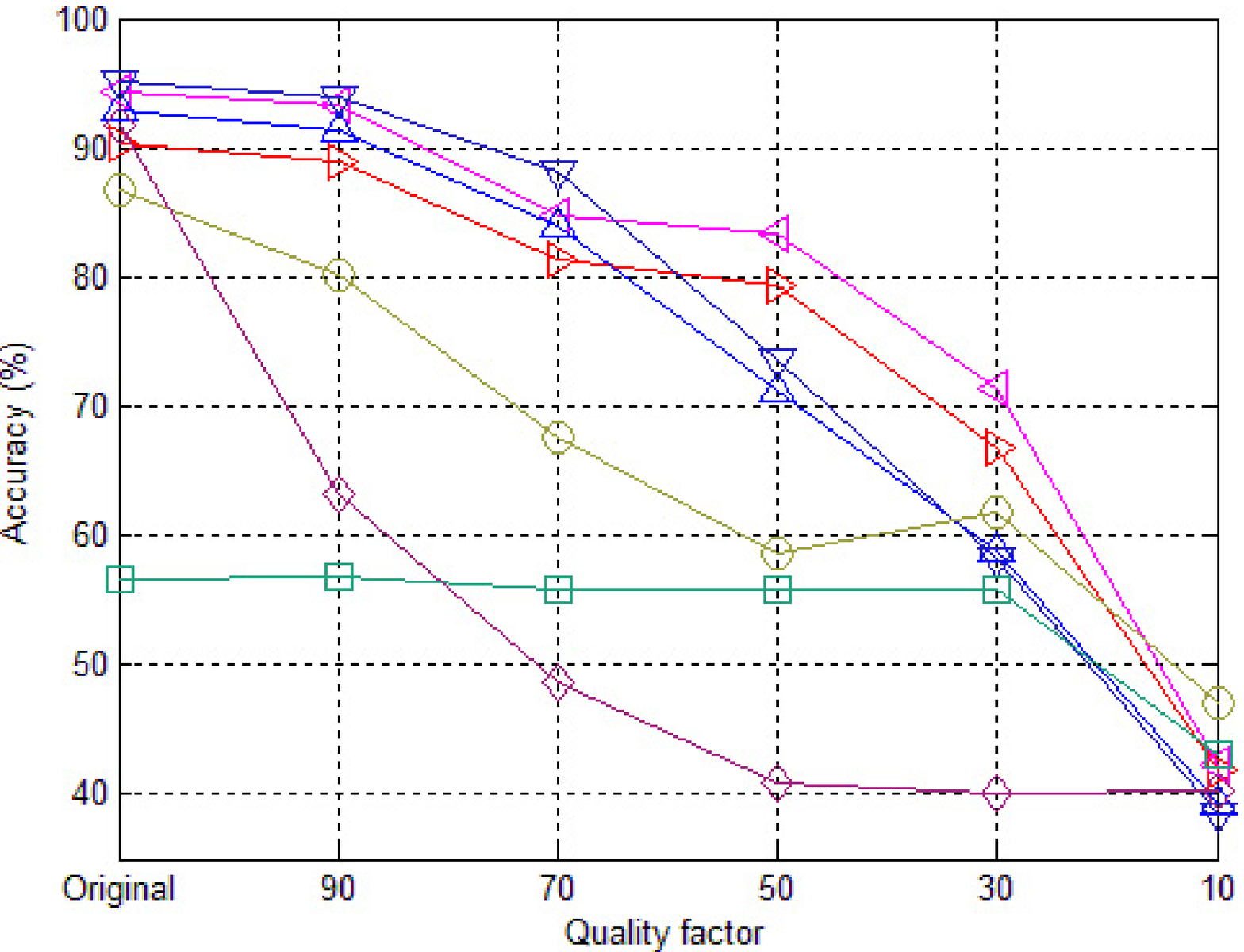}}
    \hspace{+0.05cm}
    \subfigure[Scaling]{\includegraphics[height=3.28cm]{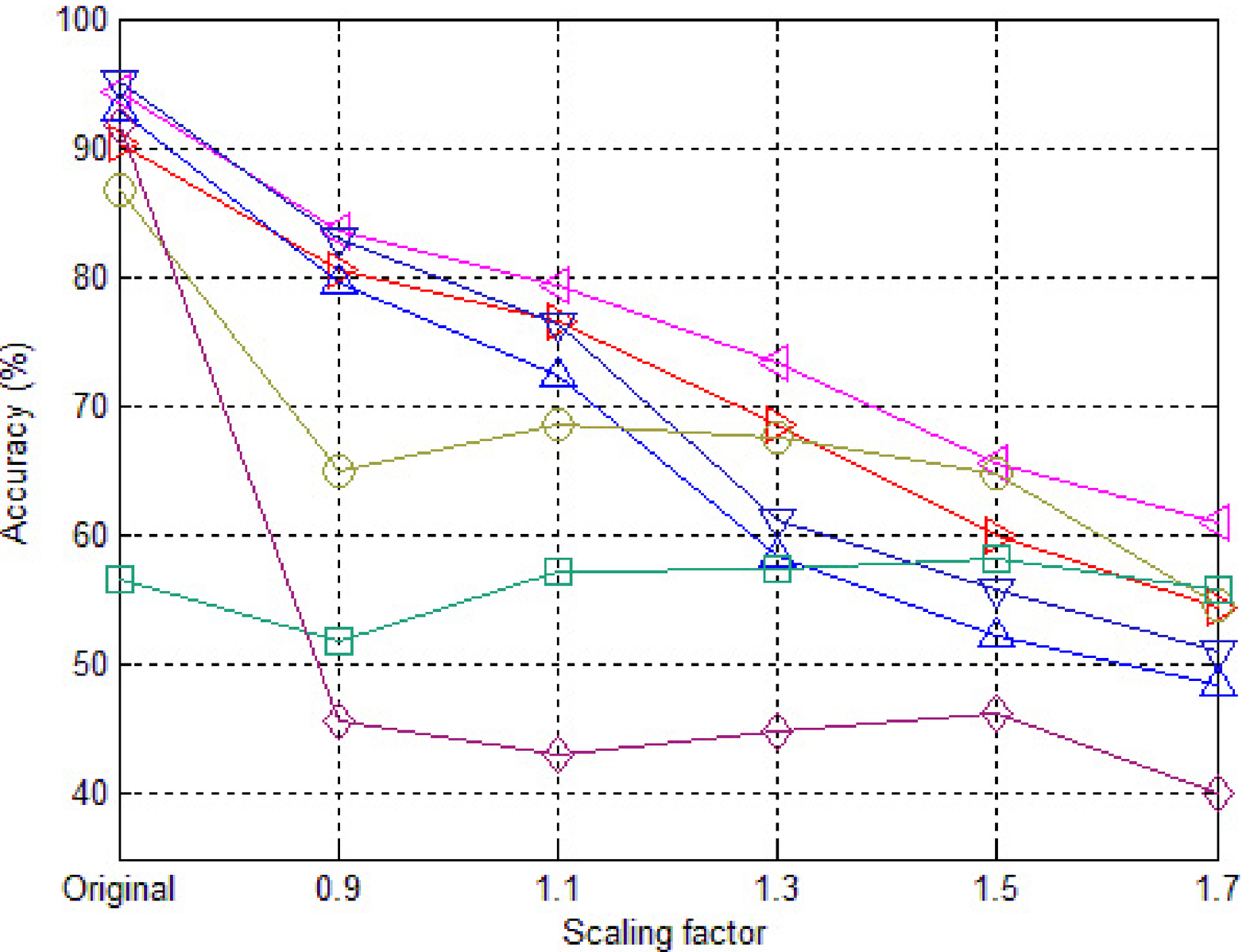}}
    \hspace{+0.05cm}
    \subfigure[Geometric transformation]{\includegraphics[height=3.28cm]{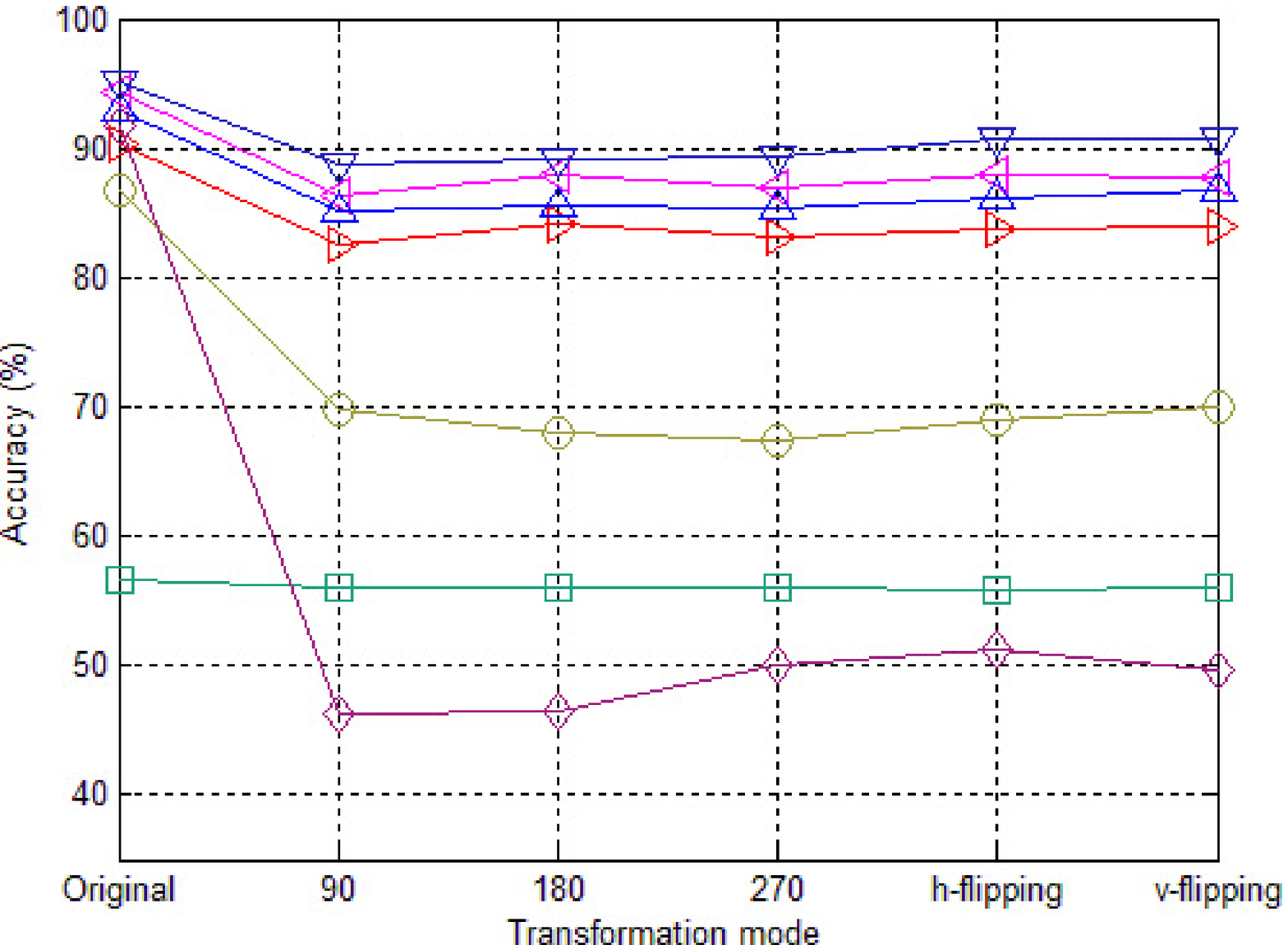}}
    \hspace{-0.1cm}
    \subfigure[Contrast stretching]{\includegraphics[height=3.28cm]{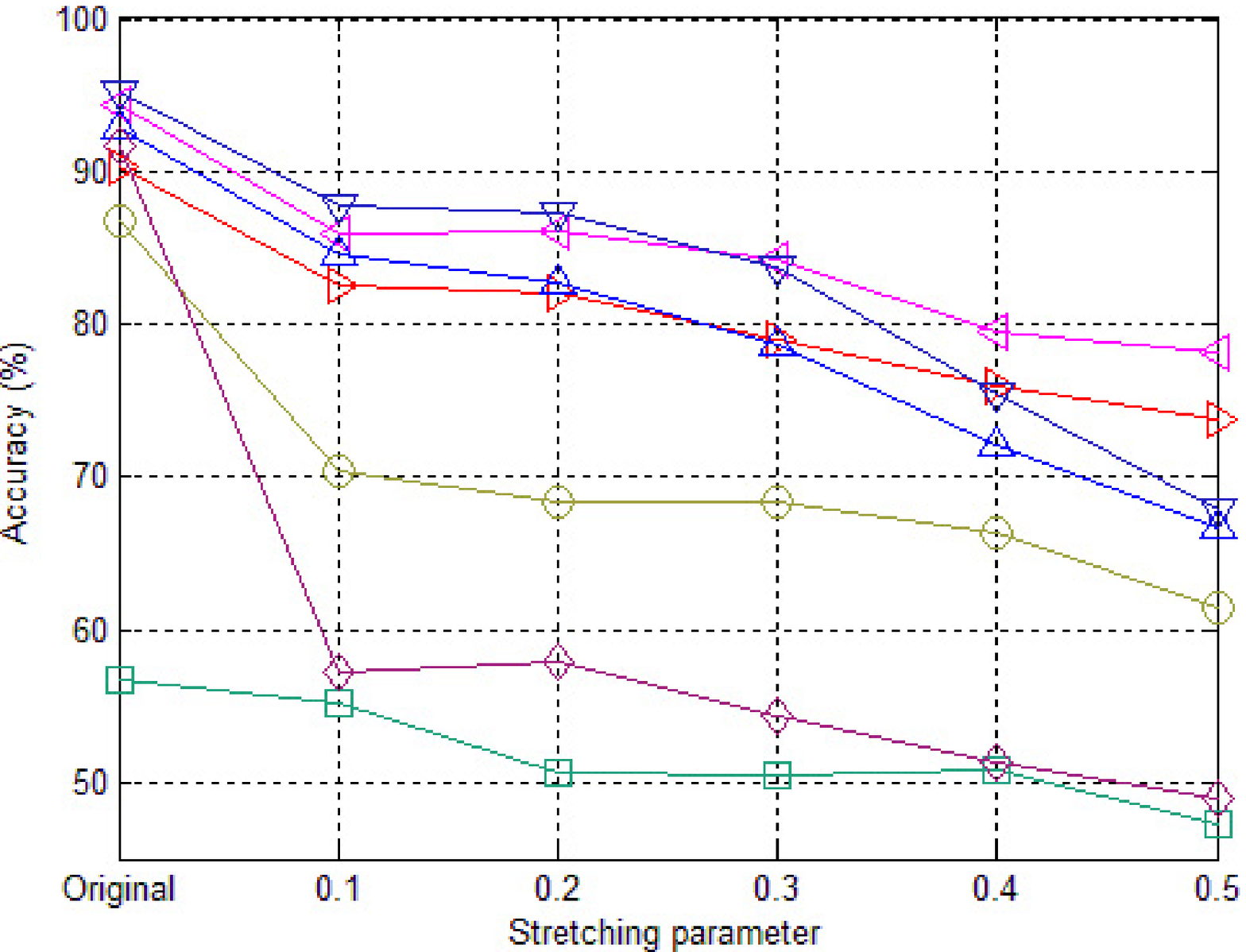}}

    \vspace{-0.3cm}
    \subfigure{\includegraphics[height=0.48cm]{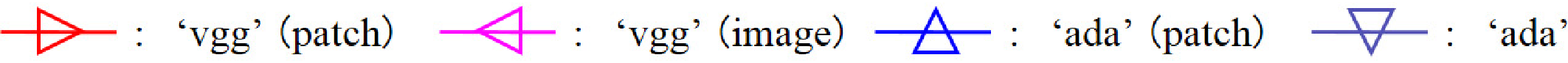}}

    \vspace{-0.3cm}
    \caption{Accuracy evolutions for testing robustness against JPEG compression, scaling, geometric transformation, and contrast stretching. Notation `h-flipping' and `v-flipping' mean horizontal and vertical flipping, respectively.}
    \label{fig:robustness}
\vspace{-0.3cm}
\end{figure*}

\begin{table}[!]
	\begin{center}

        \caption{Identification accuracy and performance comparison.}
        \vspace{-0.0cm}
        \begin{tabular}{c}
			
			\scalebox{0.22}[0.22]{\includegraphics{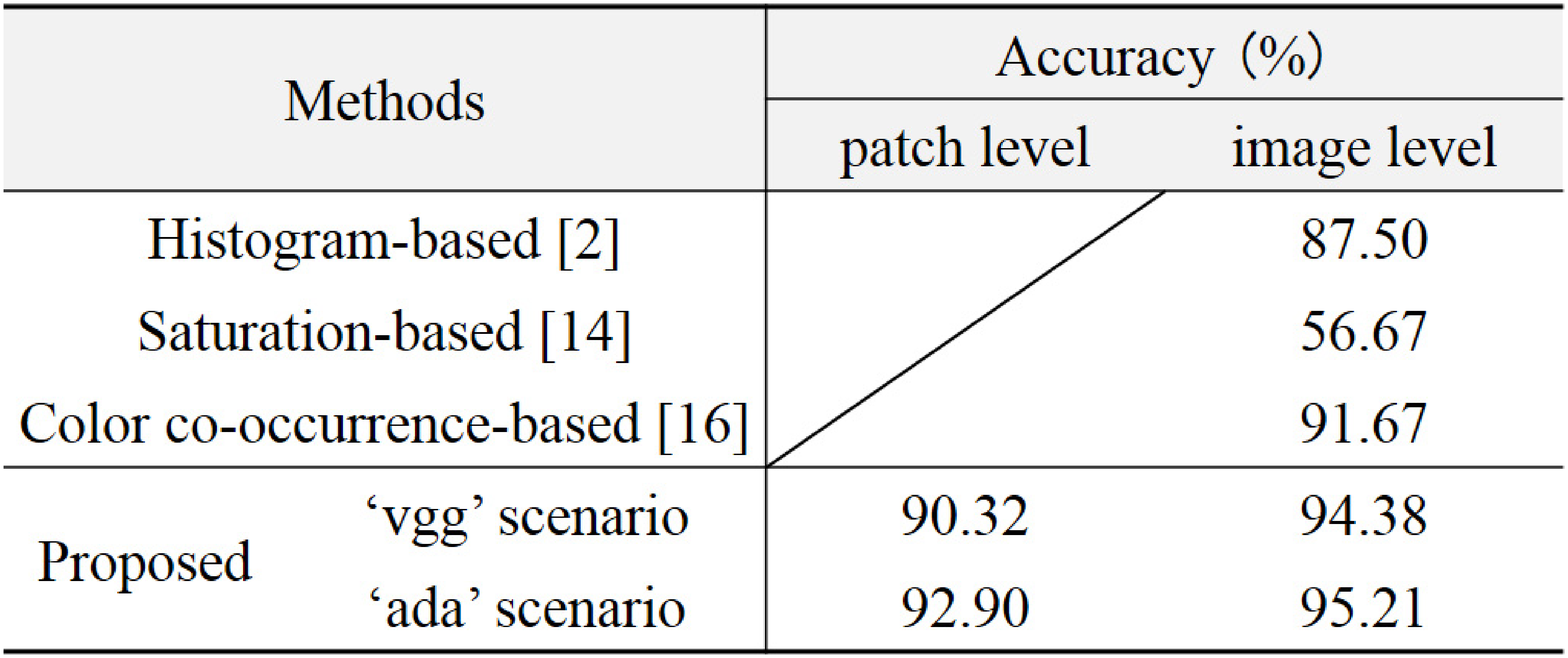}}
			
		\end{tabular}

		\label{tab:acc}
	\end{center}
\vspace{-0.7cm}
\end{table}

\begin{figure}[!]
    \centering

    \subfigure[`vgg' filters]{\includegraphics[height=3.9cm]{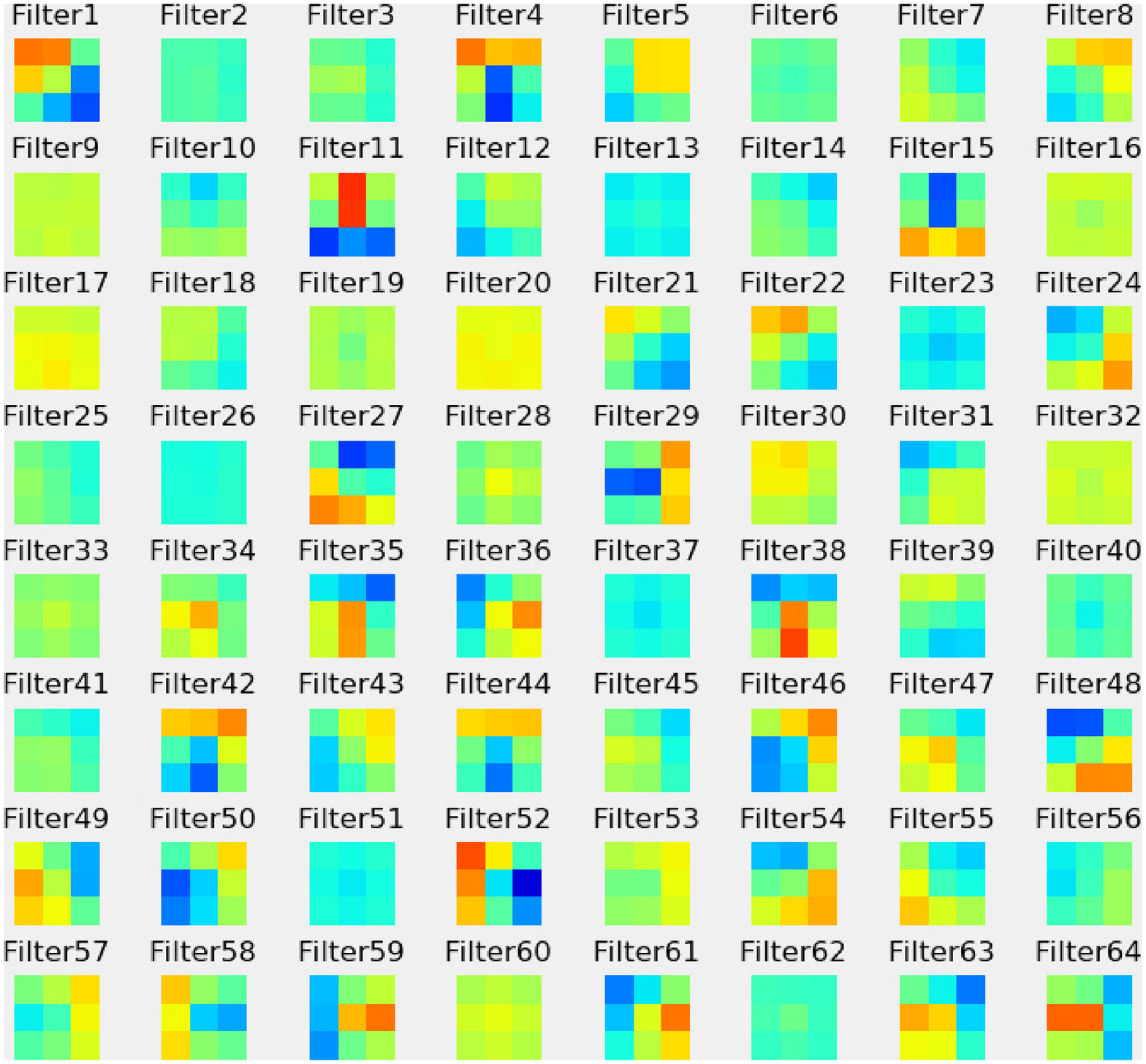}}
    \hspace{+0.0cm}
    \subfigure[`ada' filters]{\includegraphics[height=3.9cm]{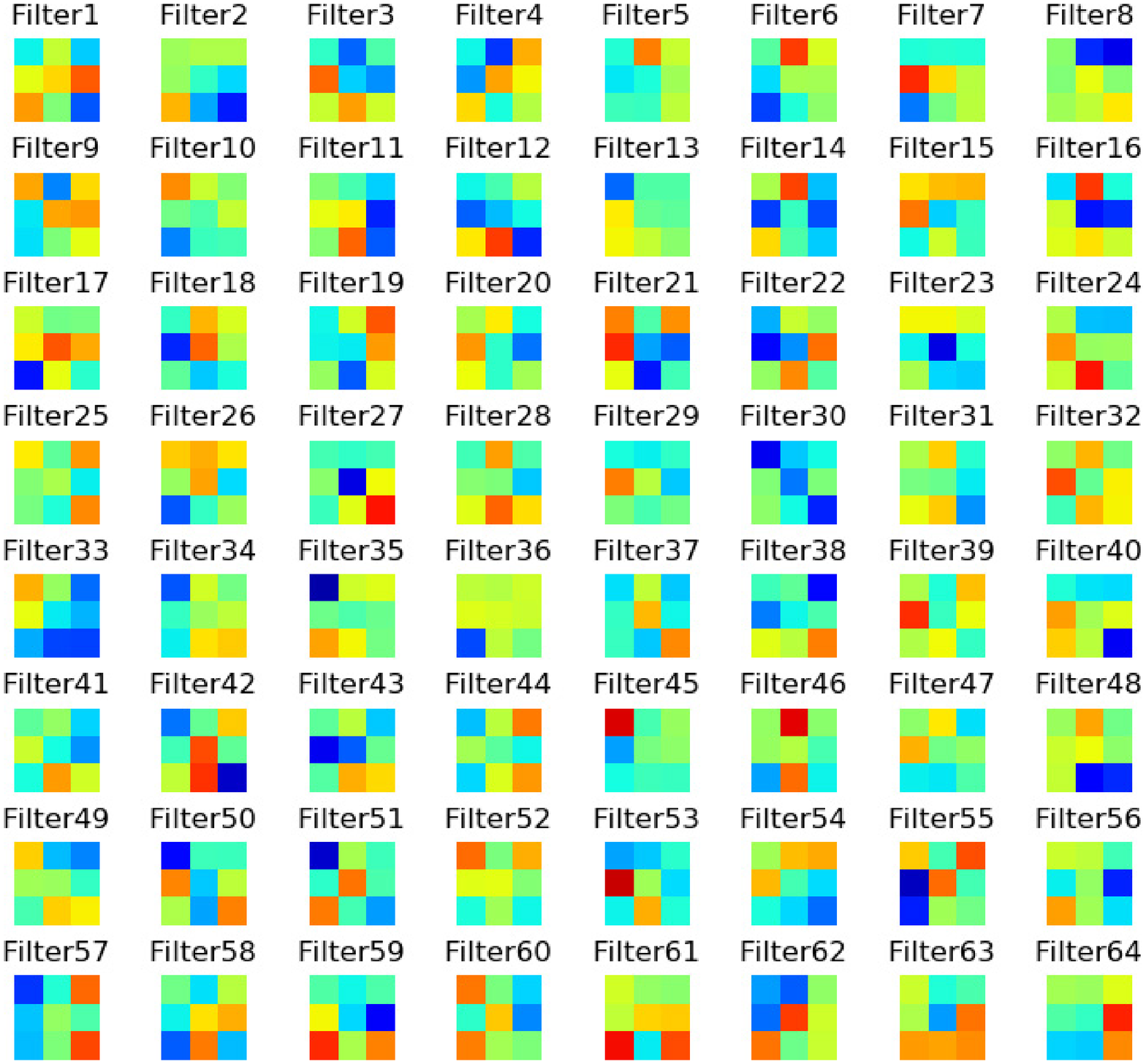}}

    \vspace{-0.2cm}
    \subfigure{\includegraphics[height=0.65cm]{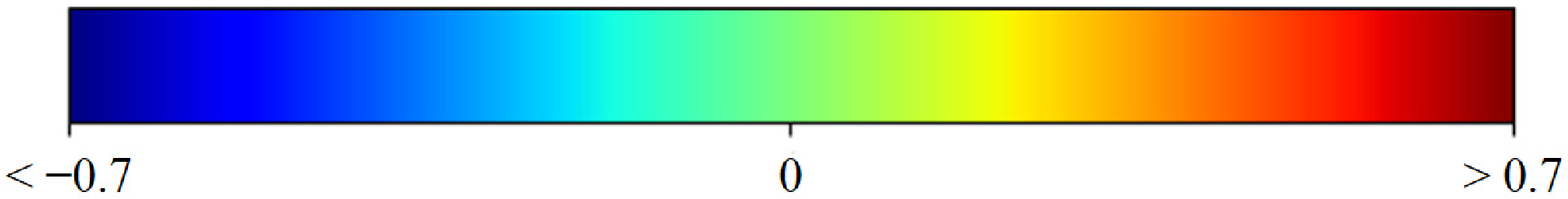}}

    \vspace{-0.3cm}
    \caption{Visualizing trained filters at the first convolutional layer (only displaying those in the third channel for space considerations).}
    \label{fig:filter_s}
\vspace{-0.5cm}
\end{figure}

\vspace{-0.06cm}
\subsection{Identification Accuracy Results}
\vspace{-0.06cm}
Figure \ref{fig:patch_r} shows three examples of origin predication at the patch level. The outputs of the CNN model are indicated by the red, green, and blue bounding boxes, which correspond to `NPI', `DGI' and `CGG', respectively. The panel above each resulting image shows the voting table, which contains the number of bounding boxes by color. As we saw, the majority voting stage has the ability to suppress the objectionable outliers, thereby improving identification performance at the image level.
\par The identification accuracy results obtained using the test set are summarized in Table \ref{tab:acc}. Patch-level accuracy is defined as the ratio of the number of correct predictions to the total number of test patches while image-level accuracy is defined as the ratio of the number of correct identifications to the total number of test images. The proposed CNN-based method identified image's origin with $95.21\%$ accuracy for the `ada' scenario. The results in Table \ref{tab:acc} show that our method has higher accuracy scores than the three baseline methods. In addition, the majority voting indeed boosted identification accuracy by $4.06\%$ for the `vgg' scenario and $2.31\%$ for the `ada' scenario.

\vspace{-0.06cm}
\subsection{Frequency Components Results}
\vspace{-0.06cm}
To study which frequency components contain information about the origin forensic, we focused on the first layer's filters, which had been trained under the `vgg' or `ada' scenario. Figure \ref{fig:filter_s} shows images of the 64 trained filters in the third channel. We found that the `ada' scenario forced the filter weights to diversify, meaning that more high-frequency components were extracted and that more image contents were suppressed. In contrast, the `vgg' filters tended to extract richer frequency components. This phenomenon is also supported, in Table \ref{tab:var}, by the channel-wise variances of the filter weights in the sense that the variances for the `ada' scenario were about twice those for the `vgg' scenario.
\par As shown by the accuracy scores in Table \ref{tab:acc}, the `ada' scenario outperformed the `vgg' one, indicating that high-frequency components possess more clues for identifying the origin. We also found that several `ada' filters, like the $47^{\text{th}}$ one, were devoted to middle-frequency components. This suggests that the middle-frequency components also contribute to the origin identification.

\vspace{-0.06cm}
\subsection{Robustness Results}
\vspace{-0.06cm}
Since artificially created images may be further processed using common image processing operations for special purposes, robustness against various post-processing operations is highly desirable for a good identification system. We thus examined the proposed method in terms of robustness against JPEG compression, scaling, geometric transformation, and contrast stretching. To this end, we prepared new test sets. For JPEG robustness, each test image was compressed for a large range of quality factors from 90 to 10 with a step size of 20. For scaling robustness, we used bicubic interpolation to resize the test images with scaling factors varying in the range [0.9,1.7]. The geometric transformations included rotation ($90^{\circ}$, $180^{\circ}$, $270^{\circ}$), and horizontal/vertical flipping. For robustness against contrast stretching, we used a piecewise-linear function that expanded the range of intensity levels so as to span the full intensity range. Two locations of points $(r_\text{min},0)$ and $(r_\text{max},255)$ control the shape of the function, where $r_\text{min}$ and $r_\text{max}$ denote the minimum and maximum intensity levels in the image, respectively. We designed a stretching parameter, $\alpha$, and manipulated $r_\text{min}$ and $r_\text{max}$: $r'_\text{min} = (1+\alpha)\cdot r_\text{min}$, and $r'_\text{max} = (1-\alpha)\cdot r_\text{max}$. Note that patch cropping was triggered again for each post-processed test image.

\par We conducted the robustness comparison with the three baseline methods. The results were plotted in Fig. \ref{fig:robustness}, where `Original' indicates the accuracy scores listed in Table \ref{tab:acc}. As shown in Fig. \ref{fig:robustness} (c), the trained CNN model was particularly robust against geometric transformations. This robustness is attributed to the diversity of the training set after data augmentation. Moreover, the `vgg' scenario achieved a comeback win over the `ada' one when the test images were severely processed. This is because the `vgg' filters, which were trained using a large-scale data set, had better generalization to deal with diverse samples.

\begin{table}[t]
	\begin{center}

        \caption{Channel-wise weight variances of first layer.}
        \vspace{-0.0cm}
        \begin{tabular}{c}
			
			\scalebox{0.22}[0.22]{\includegraphics{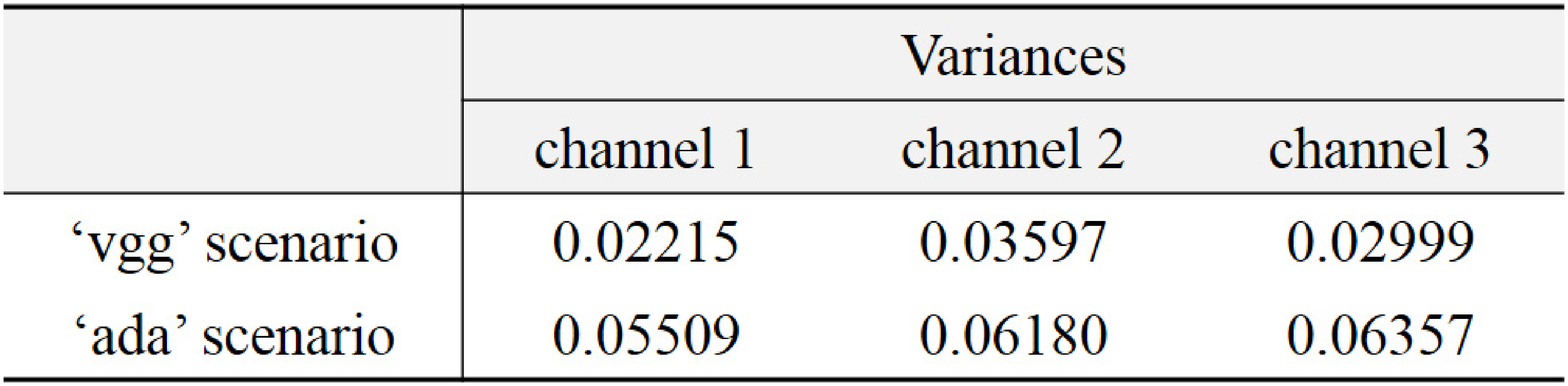}}
			
		\end{tabular}

		\label{tab:var}
	\end{center}
\vspace{-0.9cm}
\end{table}

\par For all four post-processing operations, the accuracy curves for the proposed method progressively dropped while those for the histogram-based method \cite{Wu_ICIP} and color co-occurrence-based method \cite{Li_color} dramatically deteriorated. Although the saturation-based method \cite{McCloskey} had the most stable curves, its accuracy scores were lower than $60\%$. Except for several extreme cases, our method outperformed the three baseline methods even when the test images had undergone post-processing operations. Consequently, the proposed method is better in terms of robustness than the other methods \cite{Wu_ICIP,McCloskey,Li_color}.

\vspace{-0.06cm}
\section{Conclusion}
\vspace{-0.06cm}
We have presented an effective method based on a convolutional neural network for identifying the origin of digital images. Roulette wheel selection is used to crop 200 patches from an image while considering edge information. After data augmentation, the CNN is trained to predict the origin at the patch level while the origin of the full-size image is determined by majority voting. We designed two training scenarios, namely the `vgg' and the `ada'. The experimental results show that the `ada' filters densely extracted high-frequency features and achieved the highest accuracy $95.21\%$. The `vgg' filter, however, tended to probe various frequency components and exhibited stronger robustness. Comparison of the results with those of three handcrafted feature-based methods demonstrated that the proposed method is better in terms of identification accuracy and robustness.

% Below is an example of how to insert images. Delete the ``\vspace'' line,
% uncomment the preceding line ``\centerline...'' and replace ``imageX.ps''
% with a suitable PostScript file name.
% -------------------------------------------------------------------------

%\vfill
%\pagebreak

\vfill\pagebreak

%\bibliographystyle{IEEEbib}
%\bibliography{strings,refs}

\begin{thebibliography}{00}
\bibitem{Lyu_TSP} S. Lyu, H. Farid, ``How realistic is photorealistic,'' \emph{IEEE Trans. Signal Process.}, 53(2):845-850, 2005.

\bibitem{Wu_ICIP} R.Y. Wu, X.L. Li, B. Yang, ``Identifying computer generated graphics via histogram features,'' in \emph{Proc. $18^{th}$ IEEE Int. Conf. Image Process.}, pp.1933-1936, 2011.

\bibitem{Zhang_IWDW} R. Zhang, R.D. Wang, T.T. Ng, ``Distinguishing photographic images and photorealistic computer graphics using visual vocabulary on local image edges,'' in \emph{Proc. $10^{th}$ Int. Workshop Digit. Watermarking}, pp.292-305, 2011.

\bibitem{Wang_CVIU} X.F. Wang, Y. Liu, B.C. Xu, L. Li, J.R. Xue, ``A statistical feature based approach to distinguish PRCG from photographs,'' \emph{Comput. Vis. Image Underst.}, 128:84-93, 2014.

\bibitem{Nguyen_IWDW} H.H. Nguyen, H.Q. Nguyen-Son, T.D. Nguyen, I. Echizen, ``Discriminating between computer-generated facial images and natural ones using smoothness property and local entropy,'' in \emph{Proc. $14^{th}$ Int. Workshop Digit. Watermarking}, pp.39-50, 2015.

\bibitem{Peng_IJEC} F. Peng, D.L. Zhou, M. Long, X.M Sun, ``Discrimination of natural images and computer generated graphics based on multi-fractal and regression analysis,'' \emph{AEU-Int. J. Electron. Commun.}, 71:72-81, 2017.

\bibitem{Bianchi_TIFS} T. Bianchi, A. Piva, ``Image forgery localization via block-grained analysis of JPEG artifacts,'' \emph {IEEE Trans. Inf. Forensics Secur.}, 7(3):1003-1017, 2012.

\bibitem{Ferrara_TIFS} P. Ferrara, T. Bianchi, A.D. Rosa, A. Piva, ``Image forgery localization via fine-grained analysis of CFA artifacts,'' \emph {IEEE Trans. on Inf. Forensics Secur.}, 7(5):1566-1577, 2012.

\bibitem{Yerushalmy_IJCV} I. Yerushalmy, H. Hel-Or, ``Digital image forgery detection based on lens and sensor aberration,'' \emph {Int. J. Comput. Vis.}, 92(1):71-91, 2011.

\bibitem{Chen_TIFS} M. Chen, J. Fridrich, M. Goljan, J. Luk\`{a}\v{s}, ``Determining image origin and integrity using sensor noise,'' \emph {IEEE Trans. Inf. Forensics Secur.}, 3(1):74-90, 2008.

%\bibitem{Lyu_IJCV} S.W. Lyu, X.Y. Pan, and X. Zhang, ``Exposing region splicing forgeries with blind local noise estimation,'' \emph {International Journal of Computer Vision}, vol.110, no.2, pp.202-221, 2014.

%\bibitem{TChen_TIP} T. Chen, S.J. Lu, and J.Y. Fan, ``SS-HCNN:semi-supervised hierarchical convolutional neural network for image classification,'' \emph{IEEE Transactions on Image Processing}, vol.28, no.5, pp.2389-2398, 2019.

%\bibitem{Niu_TIP} Y.L. Niu, Z.W. Lu, J.R. Wen, T. Xiang, S.F. Chang, ``Multi-modal multi-scale deep learning for large-scale image annotation,'' \emph{IEEE Trans. Image Process.}, 28(4):1720-1731, 2019.

%\bibitem{HChen_TIP} H. Chen, Y.F. Li, ``Three-stream attention-aware network for RGB-D salient object detection,'' \emph{IEEE Trans. Image Process.}, 28(6):2825-2835, 2019.

\bibitem{Rahmouni_WIFS} N. Rahmouni, V. Nozick, J. Yamagishi, I. Echizen, ``Distinguishing computer graphics from natural images using convolution neural networks,'' in \emph{Proc. IEEE Int. Workshop Inf. Forensics Secur.}, pp.1-6, 2017.

\bibitem{Yu_ICIP} I.J. Yu, D.G. Kim, J.S. Park, J.U. Hou, S.H. Choi, H.K. Lee,  ``Identifying photorealistic computer graphics using convolutional neural networks,'' in \emph{Proc. $24^{th}$ IEEE Int. Conf. Image Process.}, pp.4093-4097, 2017.

\bibitem{Quan_TIFS} W.Z. Quan, K. Wang, D.M. Yan, X.P. Zhang, ``Distinguishing between natural and computer-generated images using convolutional neural networks,'' \emph {IEEE Trans. Inf. Forensics Secur.}, 13(11):2772-2787, 2018.

\bibitem{McCloskey} S. McCloskey, M. Albright, ``Detecting GAN-generated imagery using color cues,'' \emph{arXiv preprint:1812.08247}, 2018.

\bibitem{Marra} F. Marra, D. Gragnaniello, L. Verdoliva, G. Poggi, ``Do GANs \linebreak leave artificial fingerprints,'' \emph{arXiv preprint:1812.11842v1}, 2018.

\bibitem{Li_color} H.D. Li, B. Li, S.Q. Tan, J.W. Huang, ``Detection of deep network generated images using disparities in color components,'' \emph{arXiv preprint:1808.07276v2}, 2019.

\bibitem{Cozzolino} D. Cozzolino, J. Thies, A. R{\"o}ssler, C. Riess, ``ForensicTransfer: weakly-supervised domain adaptation for forgery detection,'' \emph{arXiv preprint:1812.02510}, 2018.

\bibitem{Yu} N. Yu, L. Davis, M. Fritz, ``Attributing fake images to GANs: analyzing fingerprints in generated images,'' \emph{arXiv preprint:1811.08180}, 2018.

\bibitem{Meso} D. Afchar, V. Nozick, J. Yamagishi, I. Echizen, ``MesoNet: a compact facial video forgery detection network,'' \emph{Proc. IEEE Int. Workshop Inf. Forensics Secur.}, pp.1-6, 2018.

\bibitem{BTAS} H.H. Nguyen, F.M. Fang, J. Yamagishi, I. Echizen, ``Multi-task Learning for Detecting and Segmenting Manipulated Facial Images and Videos,'' \emph{Proc. $10^{th}$ IEEE Int. Conf. Biom. Theory, Appl. Syst. (BTAS)}, pp.1-8, 2019.

\bibitem{vgg} K. Simonyan, A. Zisserman, ``Very deep convolutional networks for large-scale image recognition,'' in \emph{Proc. $32^{nd}$ Int. Conf.  Mach. Learn.}, pp.1-14, 2015.

\bibitem{adam} D.P. Kingma, J.L. Ba, ``ADAMM: a method for stochastic optimization,'' in \emph{Proc. $3^{rd}$ Int. Conf. Learn. Represent.}, pp.1-15, 2015.

\bibitem{dropout} N. Srivastava, G. Hinton, A. Krizhevsky, I. Sutskever, and R. Salakhutdinov, ``Dropout: a simple way to prevent neural networks from overfitting,'' \emph{J. Mach. Learn. Res.}, 15(1):1929-1958, 2014.

\bibitem{dataset1} T.T. Ng, S.F. Chang, J. Hsu, and M. Pepeljugoski, ``Columbia \linebreak photographic images and photorealistic computer graphics dataset,'' \emph{ADVENT Tech. Rep. \#205-2004-5, Columbia University}, 2005.

\bibitem{proGAN} T. Karras, T. Aila, S. Laine, J. Lehtinen, ``Progressive growing of GANs for improved quality, stability, and variation,'' \emph{Proc. $6^{th}$ Int. Conf. Learn. Represent.}, pp.1-26, 2018.

\bibitem{dataset2} F. Yu, Y.D. Zhang, S.R. Song, A. Seff, J.X. Xiao, ``LSUN: Construction of a large-scale image dataset using deep learning with humans in the loop,'' \emph{arXiv preprint:1506.03365}, 2015.

\bibitem{dataset3} Z.W. Liu, P. Luo, X.G. Wang, X.O. Tang, ``Deep learning face attributes in the wild,'' \emph{IEEE Int. Conf. on Comput. Vis.}, pp.3730-3738, 2015.

\end{thebibliography}

\end{document}